\documentclass{article}

\usepackage{arxiv}
\usepackage[utf8]{inputenc} 
\usepackage[T1]{fontenc}    
\usepackage{hyperref}       
\usepackage{url}            
\usepackage{booktabs}       
\usepackage{amsfonts}       
\usepackage{nicefrac}       
\usepackage{microtype}      
\usepackage{lipsum}
\usepackage{graphicx}
\usepackage{subcaption}     
\usepackage{multirow}       
\usepackage{longtable}      
\graphicspath{ {./images/} }

\usepackage[backend=bibtex8,bibencoding=ascii,style=numeric-comp,sorting=none]{biblatex}
\title{Detecting spells in fantasy literature with a transformer based artificial intelligence}

\author{
 Marcel Moravek \\
  Computer Science\\
  Hochschule Darmstadt\\ 
  University of Applied Sciences\\
  64295 Darmstadt, Germany \\
  \texttt{marcel.moravek@yahoo.de} \\
   \And
 Alexander Zender \\
  Computer Science\\
  Hochschule Darmstadt\\ 
  University of Applied Sciences\\
  64295 Darmstadt, Germany \\
  \texttt{alexander.zender@h-da.de} \\
  \And
 Andreas M\"uller\\
  Computer Science\\
  Hochschule Darmstadt\\
  University of Applied Sciences\\
  64295 Darmstadt, Germany \\
  \texttt{andreas.mueller@h-da.de} \\
}

\date{June 2022}

\begin{document}

\maketitle
\begin{abstract}
Transformer architectures and models have made significant pro\-gress in language-based tasks. 
In this area, is BERT one of the most widely used and freely available transformer architecture.
In our work, we use BERT for context-based phrase recognition of magic spells in the Harry Potter novel series. 
Spells are a common part of active magic in fantasy novels. Typically, spells are used in a specific context to achieve a supernatural effect.
A series of investigations were conducted to see if a Transformer architecture could recognize such phrases based on their context in the Harry Potter saga.
For our studies a pre-trained BERT model was used and fine-tuned utilising different datasets and training methods to identify the searched context. 
By considering different approaches for sequence classification as well as token classification, it is shown that the context of spells can be recognised.
According to our investigations, the examined sequence length for fine-tuning and validation of the model plays a significant role in context recognition. 
Based on this, we have investigated whether spells have overarching properties that allow a transfer of the neural network models to other fantasy universes as well.
The application of our model showed promising results and is worth to be deepened in subsequent studies.
\end{abstract}

\section{Introduction}
\label{introduction}
Recent developments in natural language processing have made significant advances in a wide range of tasks, such as reading comprehension \cite{wang2018}, fill-in-the-blank tasks \cite{mnli} or question-and-answer tasks \cite{squad2}. One of the currently most discussed large language models is ChatGPT\cite{openai2022chatgpt}, which has recently given a boost to the topic of artificial intelligence. ChatCPT is primarily the actual frontend to the GPT-4 model of AI. GPT-4 is a Generative Pre-trained Transformer model \cite{openai2023gpt4} which is pre-trained to predict the next token in a document. It allows https://www.overleaf.com/project/63e6b62818ebddeca5ecd3c2to conduct question and response dialogues in human-like natural language and to assign tasks in everyday language \cite{openai2023gpt4}.
In particular, the development of various transformer-based models has contributed significantly to the progress of machine language understanding in recent years \cite{overview}. The transformer \cite{transformer} is a novel model architecture that aims to solve sequence-to-sequence tasks while making the processing of long-range dependencies more efficient. 
Self-attention layers are the essential feature of transformer architectures. 
This concept of self-attention \cite{attention} allows transformer models to look bidirectionally at all words in an input sequence to better understand the context of a sequence. Moreover, self-attention is not applied once but several times in the transformer architecture, in parallel and independently \cite{transformer}.
There exist several flavours of transformer architectures, some of them use only encoder stacks as BERT, others utilise only decoder stacks as GPT-4. Originally mixed models that make use of both of them, encoders and decoders, where used for sentence translation tasks by Vaswani et al. \cite{transformer}.

However, it is still a challenge to correctly classify the context of sequences, as the context usually consists of more than just single terms \cite{lampert2021}. 
Whereas sentiment analysis is a common task for neural networks, more sophisticated linguistic constructs are ways more difficult to identify. Sarcasm, irony, jokes or spells fall in the latter categories.
Especially spells from fantasy novels present an interesting challenge that is quite complex in this field, but still rather rarely considered. Since the beginning of the 21st century, the fantasy genre has become a relevant part of contemporary literature \cite{rothfuss, rules}.
In fantasy literature magic spells are used to create a mystical and supernatural effect \cite{lawsAndFunctions}. Especially this use of magic is the essential characteristic that distinguishes fantasy from other genres \cite{nikolajeva, saricks2009readers, sanderson}. In the fantasy genre, magic generally refers to the existence and use of a power that does not exist within real natural laws. Nevertheless it typically exists within a distinct system with more or less defined and fixed rules \cite{dobby}.

\section{Related Work}
\label{related_work}
Since the publication of the first Harry Potter novel 20 years ago, a total of seven volumes have been published, with the last volume divided into two parts. The author J.K. Rowling has created an entire fantasy universe consisting of a series of specific names, places, and spells. In addition to her imaginative use of language to formulate names and terms, a variety of spells are already the subject of research and part of a scientific discourse. For example, G. Kumagai \cite{analysingSpells} carried out a sound-symbolic analysis in terms of syllable lengths, the number of voiced obstruents and stressed low vowels of the spells. There are also papers dealing with the linguistic origin and structural features \cite{linguaculturalRole, morphologicalHPStudy} or the literary challenge of translating \cite{languageOfMagic} the spells.
Vilares and Gómez-Rodríguez \cite{actionPrediction} explore the challenge of predicting actions from textual descriptions of scenes. Their analysis used a variety of texts from the Harry Potter domain, to predict upcoming actions. They analysed different models, such as an LSTM-based or a logistic regression approach, to infer which spell will be cast next, given a fragment of a story. 
Recent successes in sequence classification using a transformer approach are also noteworthy. A. Wadhawan et al. used transformer-based models to classify the information content of Twitter tweets related to coronavirus, achieving a successful F1 score of 0.9037 \cite{covidClassification}. 
Furthermore, A. Pritzkau, S. Winandy and T. Krumbiegel \cite{informationClassification} also used the BERT model to indicate the trustworthiness of information in tweets. In several experiments with different parameters, they achieved an F1 score of more than 0.9 several times. Therefore, the transformer architecture seems to be very suitable for our classification of phrases with or without spells. 

\begin{figure*}[h]
    \begin{subfigure}{0.5\textwidth}
        \includegraphics[width=\linewidth]{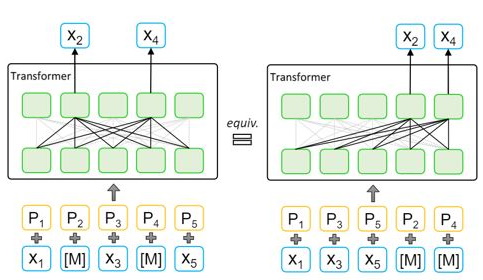}
        \caption{Masked Language Modeling (MLM)}
        \label{fig:mpnet_mlm}
    \end{subfigure}
    \begin{subfigure}{0.5\textwidth}
        \includegraphics[width=\linewidth]{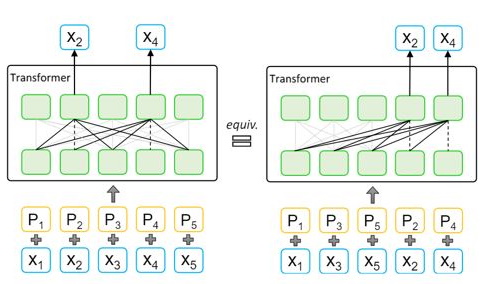}
        \caption{Permuted Language Modeling (PLM)}
        \label{fig:mpnet_plm}
    \end{subfigure}
\caption{Visualization of pre-training concepts MLM and PLM according to \cite{song2020}}
\label{fig:pre-training_concepts}
\end{figure*}

Given the large number of excellent pre-trained transformer models available, the focus of this research is more about determining a good fine-tuning approach rather than training a complete model from scratch.
One approach that has improved the performance of the original BERT models is MPNET \cite{mpnet}. Here the model architecture itself remains the same, only the procedures during training change. MPNET removes the "Next Sentence Prediction" (NSP) approach \cite{xlnet, roberta, spanBERT} and extends the "Masked Language Modeling" (MLM) by a modified version of the autoregressive pre-training method "Permutation Language Modeling" (PLM) \cite{xlnet}. 
However, by combining MLM and PLM into a unified approach, it reduces some respective limitations of the original BERT training process, such as the lack of dependencies of the masked tokens themselves \cite{transformerXL} and the unidirectional view of the original PLM approach. 
This new training approach is called masked and permuted language modelling (MPNet for short) \cite{mpnet}. The tokens of a sequence are rearranged and divided into unpredicted and predicted groups, as shown in figure \ref{fig:pre-training_concepts}. The approach takes into account the dependencies between predicted tokens through permuted language modelling, thus avoiding the MLM problem of BERT. This is achieved by first permuting the input sequence, as shown on the left side of the figure, and then shifting the masked tokens completely to the right. 
In this unified view, the non-masked tokens are placed on the left side of the input, while the masked tokens and the tokens for both MLM and PLM are placed on the right side of the permuted sequence. In figure \ref{fig:mpnet_mlm}, this is shown by moving the masked tokens X2 and X4 of the permuted sequence (X1, X3, X5, X2, X4) to the right. By conserving the position information in this way, the model will always know the correct order of the sentence and the position error is reduced.
Using this training approach with the same BERT model architecture as well as the enhanced RoBERTa \cite{roberta} training data achieves a better result on a variety of downstream NLP tasks \cite{mpnet}. The main advantage of MPNet is that the masked token prediction model incorporates more information for prediction, resulting in better output embeddings and less mismatch with downstream tasks \cite{mpnet}.

Another advance is the Sentence-BERT (SBERT) model by Reimers and Gurevych \cite{sbert}. Typically, finding the "Semantic Textual Similarity" (STS) between just a few thousand sequences requires dozens of hours of computation. The standard CLS token, providing an aggregated vector representation of the entire input, is rather unsuitable for computing sentence pair regression tasks such as STS. It leads to a performance below of average GloVe embeddings \cite{glove}. 
Extension of the BERT architecture with a siamese or triplet network structure lead to better results \cite{faceNet}. To derive the sentence embeddings, Reimers and Gurevych \cite{sbert} add a pooling operation to the output of BERT. This updates the parameters of the model so that the generated sentence embeddings can be used to compare sentences semantically. A similarity metric such as cosine similarity \cite{cosinSimilarity} or Manhattan/Euclidean distance \cite{euclidean} was used as quality metric. 
However, the similarity measures can be performed very efficiently on these embeddings. This allows to use SBERT for semantic similarity search as well as for clustering \cite{sbert}. 
To create the embeddings, three pooling strategies were compared: (1) the standard CLS output, (2) the mean of all outputs (MEAN strategy), and (3) a maximum time course of the outputs (MAX strategy). The MEAN strategy produced the best results. This involves averaging all token embeddings and consolidating them into a unique vector space, the 'sentence vector'. This reduces significiantly the computation time for finding semantically similar sentence pairs in large datasets. The calculation time needed drops from several hours to a few seconds - improving it by three orders of magnitude \cite{sbert}.

This SBERT strategy, was used to fine tune the already trained original BERT models on a dataset of 1B sentence pairs. A contrastive learning objective was utilised in the model. Selecting a sentence from the pair, the model predicts which sentence of a set of randomly sampled other sentences in the dataset matches it best. The SBERT approach outperforms the state-of-the-art results on seven STS tasks \cite{sbert}.
However, the models trained in this way are suitable as multipurpose models for different tasks, such as semantic text similarity, semantic search or paraphrase mining. 
One of the models used was the pre-trained MPNet model, which achieved the best quality \cite{mpnet}.

\section{Methodology}
\label{chap:methodolgy}
Recognising spells using a Transformer model requires both a pre-trained model and a labelled dataset for task-specific fine-tuning and evaluation. The task for the model is not simply to memorise individual spells, but to recognise them based on their context. Therefore extending the recognition capability of spells to new ones which are not included in the training dataset.

\subsection{Determining Spells}
The corpus from the Harry Potter novel is based on seven main books of the Harry Potter saga \cite{hp1, hp2, hp3, hp4, hp5, hp6, hp7}. The original ebook versions are converted into plain text files and all non-content elements, such as headings, page numbers, forewords, etc., are removed. The corpus is then split by using books one to six as the training set and book seven as the validation set. Now, a list of spells (Appendix \ref{app:appendix}) is used as search phrases to create datasets for classification by determining whether the sequence contains a magic spell (positive) or not (negative) by a simple string comparison. To reduce some grammatical complexity, the corpus and all spells are converted to lowercase.

Separating incantations from spell names can be quite difficult, especially in the case of the Harry Potter universe. This is because a significant number of spell names are often formed by simply adding words such as "spell", "charm" or "curse" to the pronounced incantation (e.g. the incantation "reducto" and the name "reductor curse"). It was decided to also use them as search phrases to create another dataset from a combination of the incantation and the spell name as positively classified search phrases. A sequence is classified as positive if it contains either a pronounced incantation or a paraphrased spell designation.

Several phrases in the universe appear to be spells, but are more likely names for other actions. Because of their similarity to spell names, the following phrases are treated as such:

\begin{itemize}
    \item Apparate/Apparition and Disapparate/Disapparition: These terms describe a kind of travelling, similar to teleportation. They are described from the viewer's point of view: disapparition - to disappear and apparition - to appear.
    \item Occlumency und Legilimency: The term Occlumency refers to a type of meditation used to seal one's mind so that others can't enter another's mind with the magical act of Legilimency. A person who practised this art was known as a Legilimens.
\end{itemize}

Some general literary paraphrases of a certain category of spells, such as the term "defensive spells", include various spells for defensive purposes. However, since such terms do not describe a specific spell or action, they are not classified as spells or spell names. Finally, a complete list of all incantations with their corresponding names, as well as all spell names without an incantation, can be found in Appendix \ref{app:appendix}.

\subsection{Context Length}
In order to determine which context length is the most efficient, the corpus is split according to three different rules. By using different splitting schemes, the length of a sequence will vary and different context lengths could be evaluated. However, the maximum context length is limited by the maximum number of sequence tokens that the model can process. We used the following three types of splits: 

\begin{itemize}
    \item Sentence-Split: The length of a sequence for the dataset is exactly one sentence from the books. The corpus is split into sentences using the sent-tokenize function of the NLTK library\footnote{\url{https://www.nltk.org}}. 
    \item Paragraph-Split: The length of a sequence for the dataset is exactly one full paragraph from the books.
    \item Sequence-Split: Here, several sentences are concatenated until the total number of tokens of the combined sentences reaches the limit of the processable sequence length of the model. A new sequence always starts with the next sentence that has not yet been used. 
\end{itemize}

The length of the sequence context increases with each split type. However, as the corpus only contains the maximum content of the seven books, the amount of all entries of the dataset decreases with each split type. A comparison of the differently generated training and evaluation datasets is given in table \ref{table:split-types}. 

\begin{table}[h!]
\centering
\begin{tabular}{ l || l | l || l | l || l | l } 
& \multicolumn{2}{c||}{Sentence-Split} & \multicolumn{2}{c||}{Paragraph-Split} & \multicolumn{2}{c}{Sequence-Split} \\
 \hline

 & Train & Eval & Train  & Eval & Train & Eval \\ [0.5ex] 
 \hline
 total & 64.355 & 14.242 & 31.324 & 6.911 & 3.640& 805 \\ 
 \hline
 incantations only & 283 & 112 & 262 & 108 & 195 & 84 \\
 combined & 773 & 284 & 721 & 271 & 541 & 204 \\ [1ex] 
 \hline
\end{tabular}

\caption{Comparison of generated datasets.}
\label{table:split-types}
\end{table}

Two datasets are generated for each split type using incantations only and incantations and spell names (combined) respectively as categories for the search phrases. The latter produces a complete data set with a significantly larger amount of data than using invocations only.

\subsection{Evalution Procedure}
Each created dataset undergoes a procedure of three steps: 
\begin{enumerate}
    \item fine-tuning of the pre-trained model with an individually created training dataset based on the split rule used, 
    \item context prediction of the paired evaluation dataset with the fine-tuned model, 
    \item and evaluation of the prediction sequences and comparison with other methods. All calculations are performed on an AWS p2.xlarge instance\footnote{\url{https://aws.amazon.com/de/ec2/instance-types/p2/}} with an NVIDIA K80 GPU with 12 GiB of GPU memory. 
\end{enumerate}

The pre-trained BERT model is loaded with the open-Source library "Transformers"\footnote{\url{https://github.com/huggingface/transformers}} \cite{huggingface} from the HuggingFace Model-Hub\footnote{\url{https://huggingface.co/}}. 
This library implements a sophisticated training routine that will be used for further fine-tuning. Through the use of different pre-implemented classes, the classification specific head component of the model could be easily replaced. 
Due to the unequal size of the datasets between pre-training and fine-tuning, the original parameters of the transformer model are  only slightly changed during the fine-tuning process. Mainly the parameters of the head layer are adapted to the classification specific task \cite{illustratedTransformer}.
The choice of hyperparameters is based on the original BERT publication\footnote{
The optimal values of the hyperparameters are task-specific, but \cite{bert} recommends the following ranges of possible values:     
\begin{itemize}
    \item Batch Size: 16, 32
    \item Learning Rate (Adam): 5e$^{-5}$, 3e$^{-5}$, 2e$^{-5}$
    \item Number of Epochs: 2, 3, 4
\end{itemize}
}.
Unless otherwise stated, the model is retrained by five epochs with a batch size of 16 and a learning rate of 2e$^{-5}$ in all subsequent studies. In addition, a maximum sequence length of 384 tokens and a train/test split of 80/20 is used. The F1 score \cite{f1} is used to measure the quality of the models during training and evaluation.

For each dataset, the fine-tuned models will be evaluated by predicting the context of sequences for the corresponding evaluation dataset. The evaluation dataset is generated by the same split procedure as the training dataset on the seventh Harry Potter novel. The number of positively and negatively classified sequences will be used to determine the F1-score and to compare the efficiency of the models.

For a detailed comparison and analysis of specific classified sequences, the library "Transformers-Interpret"\footnote{\url{https://github.com/cdpierse/transformers-interpret}} is used. By calculating several attention values for each token, the attention distributions of the individual embeddings for an input sequence could be generated. These can be used to visualise the attention weights and to determine how much influence a single token has on the overall classification performance of a sequence. The effect is given by a weight between -1 and 1 for each token. A higher weight indicates a greater influence of the token on the corresponding classification. Based on this, a feeling for the most effective tokens is obtained. An example visualisation is shown in figure \ref{fig:interpret_study_sentence_split_false}.

\section{The Transformer Architecture}
The base model for the investigations is the all-mpnet-base-v2 transformer model\footnote{\url{https://huggingface.co/sentence-transformers/all-mpnet-base-v2}}, which is a retrained variant of the MPNet model from Microsoft \cite{mpnet} through SBERT \cite{sbert}.
The model contains 12 transformer layers, a hidden-state of 768 and 110 million parameters, like the original BERTBase model \cite{mpnet, sbert}.
The creator SBERT describes the model as a general-purpose model that can be adapted to other use cases \cite{sbert}. The output of the last transformer layer of the model contains the semantic information of the input sequence, mapped into an aggregated 768-dimensional vector space. This information can be used by an additional header layer for information retrieval, clustering or sentence similarity tasks \cite{sbert}.

\begin{figure*}[h]
    \centering
    \includegraphics{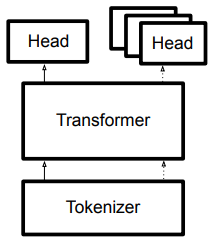}
    \caption{Building blocks of the transformer architecture}
    \label{fig:building_blocks}
\end{figure*}

The model used is provided by Hugging Face from the Transformers library. Each model in the library is defined by three building blocks, as shown in Figure \ref{fig:building_blocks} \cite{huggingface}:
\begin{itemize}
    \item A tokenizer breaks the input sequences into word chunks and converts them into an Index encoding based on the tokenizer's vocabulary.
    \item The specific architecture of the actual transformer used is implemented in the transformer module. It computes context-dependent embeddings from the index-encoded tokens.
    \item Task-specific output can be generated from the contextual embeddings using different head components. These head components are easily interchangeable and can be adapted to specific tasks.
\end{itemize}
Each model in the transformer library is represented by three components: a tokenizer, a transformer, and various interchangeable head components \cite{huggingface}.
By changing the head level, the same pre-trained model can easily be used for different tasks. The size of the data set and the amount of fine-tuning required for each task is negligible compared to pre-training with Microsoft and SBERT. During fine-tuning, mainly the parameters of the head layer are adjusted for the task, in order to produce the desired output based on the information from the generated contextual embeddings of the underlying transformer layer \cite{huggingface}.

\begin{figure}[t]
    \begin{subfigure}{0.46\textwidth}
      \includegraphics[width=\linewidth]{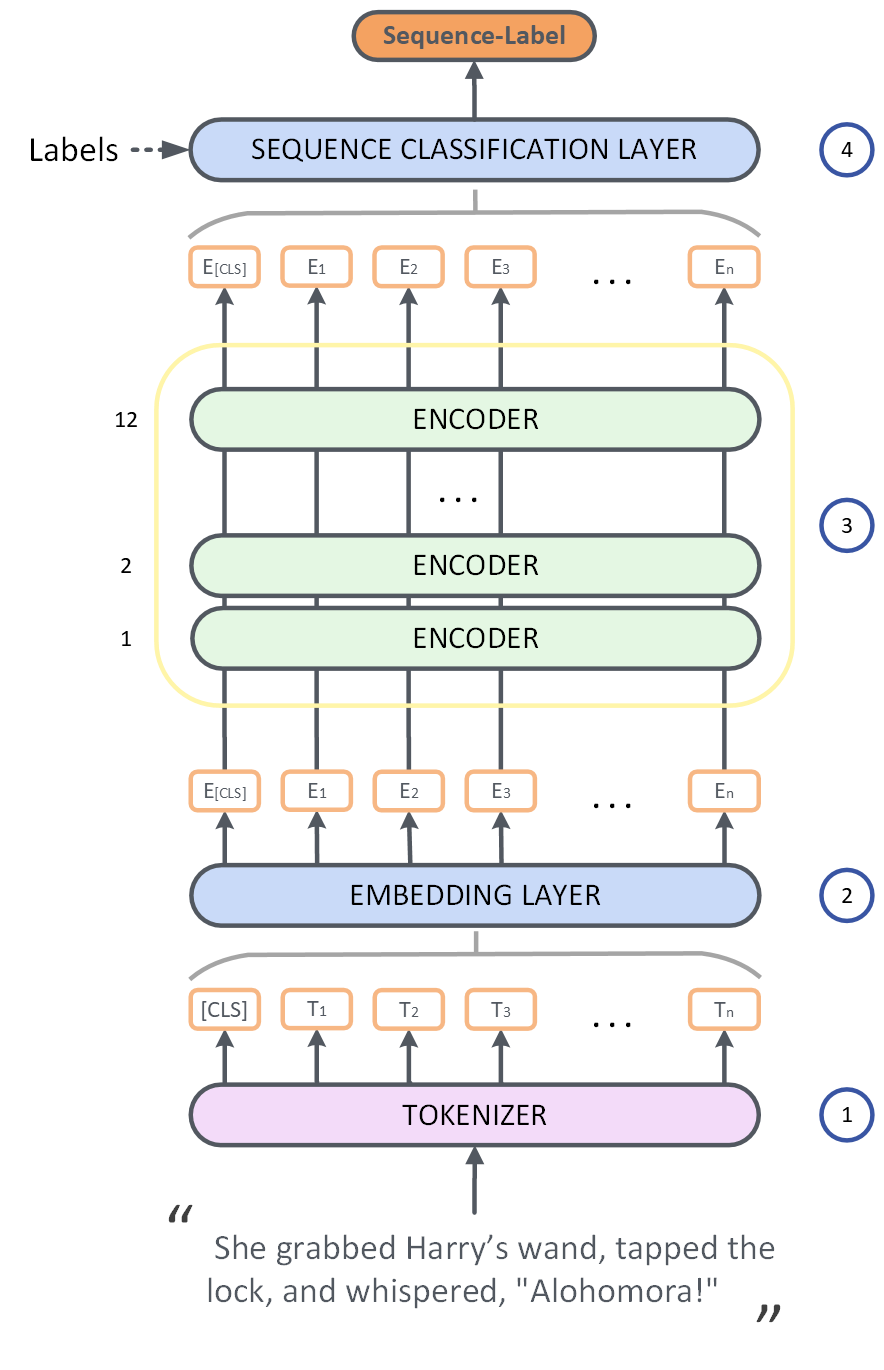}
      \caption{Sequence classification architecture}
      \label{fig:model_arch_sequence}
    \end{subfigure}
    \begin{subfigure}{0.67\textwidth}
      \includegraphics[width=\linewidth]{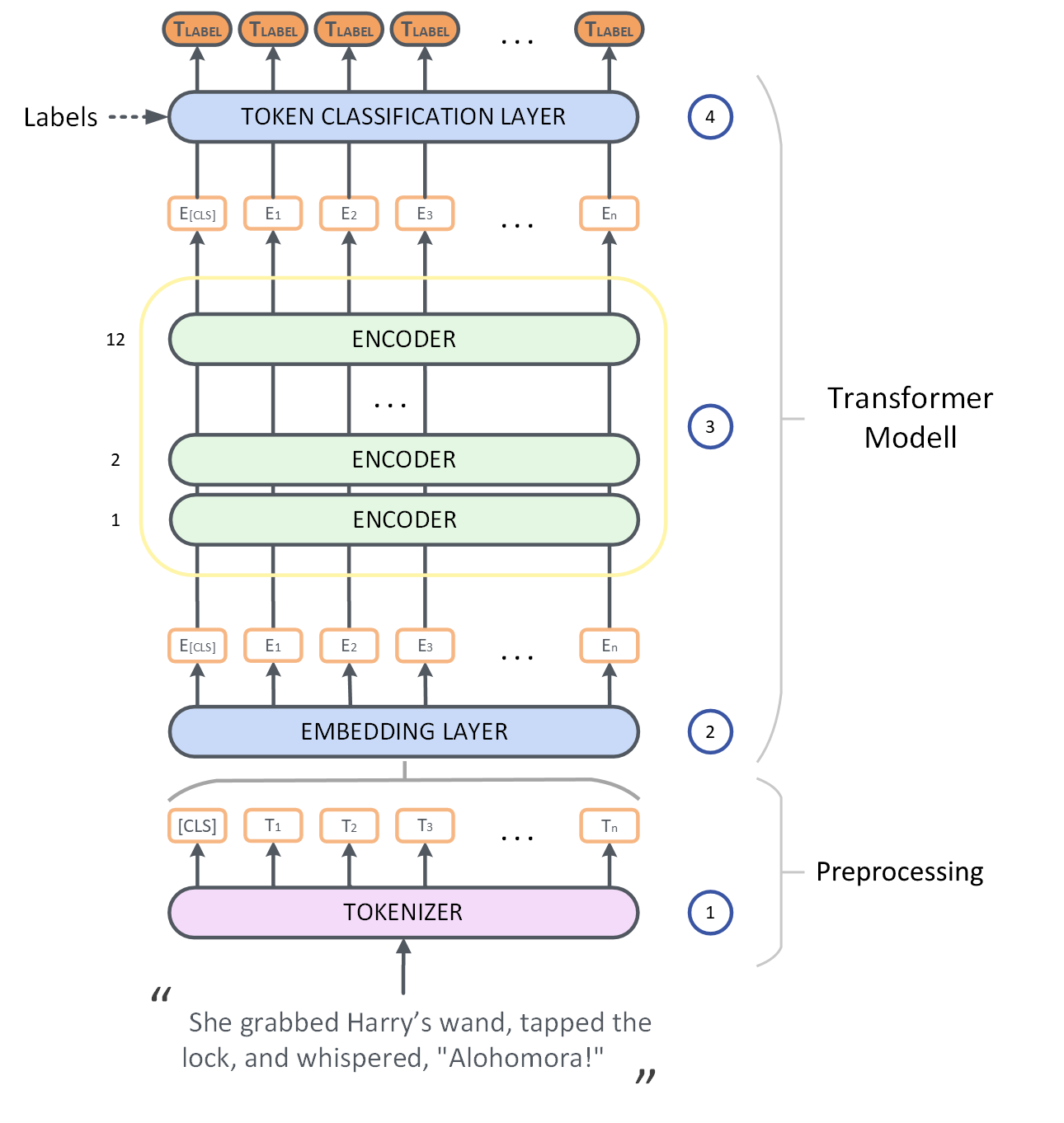}
      \caption{Token classification architecture}
      \label{fig:model_arch_token}
    \end{subfigure}
    \caption{Used model architectures}
    \label{fig:model_arch}
\end{figure}

The detailed architecture of the model is shown in Figure \ref{fig:model_arch}. 
It is divided into four components: 
\begin{enumerate}
    \item  A pre-processing step, where the input sequence is divided into tokens by a tokenizer belonging to the model with a vocabulary size of 30,522 entries. These tokens (T1 to Tn) are then transferred to the actual transformer as index coding over the vocabulary of the tokenizer.
    \item Next, the transformer model generates a vector representation (E1 to En) from the index encoding for each token. Thus, the token embeddings from the index encoding are computed with the corresponding position embeddings \cite{mpnet}. By default, the model accepts input with a sequence length of 384 tokens. Sequences with fewer tokens are padded and sequences with more tokens are truncated. 
    \item These embeddings are passed in parallel through a stack of transformer blocks to build a context reference using the self-attention layers contained in each transformer encoder block.
    \item Now different head-layers are used as seen in figure \ref{fig:model_arch_sequence} and \ref{fig:model_arch_token}. During training, the possible labels are transferred to the classification head-layer.
\end{enumerate}

For sequence classification investigations the architecture of figure \ref{fig:model_arch_sequence} is used with a linear layer that processes a pooled output label for the entire input sequence.
In addition, for more detailed investigations on token level the architecture of figure \ref{fig:model_arch_token} is used, to determine an output label for each token separately based on the output of the hidden states of the last encoder.
Each model in the Transformers library is represented by three components: a tokenizer, a transformer, and various interchangeable head components. \cite{huggingface}

Through changing the head-layer, the same pretrained model can be easily used for different tasks. The size of the dataset and the duration of the fine tuning for the individual examinations is comparatively negligible compared to the pre-trainings by microsoft and Sbert. During the fine tuning mainly the parametesr of the head-layer get adjusted for the task, to generate the desired output based on the information from the generated contextual embeddings of the underlying transformer layer \cite{huggingface}.

The detailed architecture of the model is shown in figure 4.3. It is divided into four individual components: 
(1) In a pre-processing step, the input sequence is divided into tokens with a tokenizer belonging to the model with a vocabulary size of 30,522 entries. These tokens T1 to Tn are then transferred to the actual transformer as an index encoding over the vocabulary of the tokenizer.
(2) Next the Transformer model generates a vector representation E1 to En from the index encoding for each token. Therefore the token-embeddings from the index encoding are calculated with the respective position embeddings \cite{mpnet}. By default, the model takes input with a sequence length of 384 tokens. Sequences with fewer tokens are filled using a padding and sequences with more tokens are cut off. 
(3) These embeddings are passed through a stack of transformer blocks in parallel to build up a context reference using the self-attention layers contained in the individual transformer encoder blocks.
(4) Different head layers are now used, as shown in Figure \ref{fig:model_arch_sequence} and \ref{fig:model_arch_token}. During training, the possible labels are transferred to the classification head layer.
For sequence classification investigations, the architecture of \ref{fig:model_arch_sequence} is used, with a linear layer that processes a pooled output label for the entire input sequence.
In addition, for more detailed token-level investigations, the architecture of \ref{fig:model_arch_token} is used to 
determine an output label for each token separately, based on the output of the hidden states of the last encoder.

\section{Results and Discussion}
\label{chap:results_and_discussion}
In the following, different models were created to recognise spells using both sequence classification and token classification. 
Prior to this, a simple dictionary analysis based on a string match search showed that the use of neologisms as well as the use of common terms for Harry Potter spells meant that the spells couldn't simply be recognised by their term originality. We obtained just a F1 score of 20.33 percent for the dictionary analysis. After extending the dictionary to character names the analysis still led to an unsatisfactory F1 score of 34.66 percent.

\subsection{Sequence Classification Studies}
\label{chap:sequence_classification_studies}
In sequence classification studies, the initial model is trained separately on different datasets to classify the context of a given sequence, and to find out whether a transformer can perceive the context of a spell for different context sizes.
In order to determine an appropriate context length for the sequences, various datasets were created, containing sequences with and without spells. Therefore, the corpus is divided into sequences of different context sizes using the approaches described in \ref{chap:methodolgy} to create the datasets. For the sequence classification approaches, an entry is labelled positive if the associated sequence contains a spell. Otherwise the entry is negative. The negative sequences for training are randomly selected from all negative sequences. The task for the model is to classify all entries of the evaluation dataset as positive (contains a spell) or negative (contains no spell). 

\begin{table}[ht]
\centering
\begin{tabular}{ c | l | l }
    & \multicolumn{2}{c}{\textbf{F1-Score}} \\
    \# & Kombi & Inc  \\
    \cline{1-3}
    (1 a) & 0,8705 & 0,7698 \\
    (1 b) & 0,8728 & 0,8377 \\
    (2)   & 0,9097 & 0,8945 \\
    (3 a) & 0,9563 & 0,9277 \\
    (3 b) & 0,9559 & 0,9697 \\
\end{tabular}
\caption{Comparative overview of all sequence classification studies}
\label{tab:study_sequence_overview}
\end{table}

Table \ref{tab:study_sequence_overview} shows a comparative overview of the F1-score evaluation results for all trained sequence classification models.

\subsubsection{Sentence-Split}
The models (1a) are trained on a dataset generated by the Sentence-Split approach. An entry of the dataset consists of one sentence of the corpus. In order to teach the model an understanding of the context of a spell, the literary frequency of occurrence of a spell is also observed. An attempt is made to map the proportion of sentences with and without a spell from a real novel. Therefore, the proportion of negative entries is set 10 times higher than the proportion of positive entries, resulting in a total dataset size of 3,113 entries for the dataset with spells only and 8,503 entries for the combined dataset. The number of records in the evaluation dataset is 14,242.

\begin{table}[t]
\begin{subtable}{0.48\textwidth}
    \begin{small}
        \begin{tabular}{ c l | l l}
            \multicolumn{2}{l}{} & \multicolumn{2}{c}{Predicted} \\
            &  & \textbf{Positive} & \textbf{Negative} \\
        \cline{2-4}
            \multirow{2}{*}{\rotatebox[origin=tr]{90}{True}} & \textbf{Positive} & 279 TP & 5 FN \\
        & \textbf{Negative} & 78 FP & 13880 TN \\[5pt]
        & \multicolumn{3}{l}{ F1-Sore: 0,8705} \\
        \end{tabular} 
    \end{small}
    \caption{Full Dataset} 
\label{tab:study_sentence_combined}
\end{subtable}
\hspace*{0.02\textwidth} 
\begin{subtable}{0.48\textwidth}
    \begin{small}
        \begin{tabular}{ c l | l l}
            \multicolumn{2}{l}{} & \multicolumn{2}{c}{Predicted}  \\
            &  & \textbf{Positive} & \textbf{Negative} \\
        \cline{2-4}
             \multirow{2}{*}{\rotatebox[origin=tr]{90}{True}} & \textbf{Positive} & 112 TP & 0 FN \\
            & \textbf{Negative} & 67 FP & 14.063 TN \\[5pt]
            & \multicolumn{3}{l}{F1-Sore: 0,7698} \\
        \end{tabular} 
    \end{small}
    \caption{Incantation Dataset} 
    \label{tab:study_sentence_incantations}
\end{subtable}
\caption{Metrics of trained models 1a}
\label{tab:study_sentence_split}
\end{table}

The models achieved F1 scores of 0.7698 and 0.8705. Here the model trained on incantations only is slightly worse than the model with the full training dataset. The exact confusion matrix of the results can be found in Table \ref{tab:study_sentence_split}. Looking at the falsely classified sequences, it can be seen that many words like "spell", "wand" or "charm" are often associated with a positive classification, even if they are not used within an incantation. Furthermore, descriptive effects of spells are often classified positively. The model does not succeed in clearly distinguishing between spell names and incantations.

\begin{figure}
  \includegraphics[width=\linewidth]{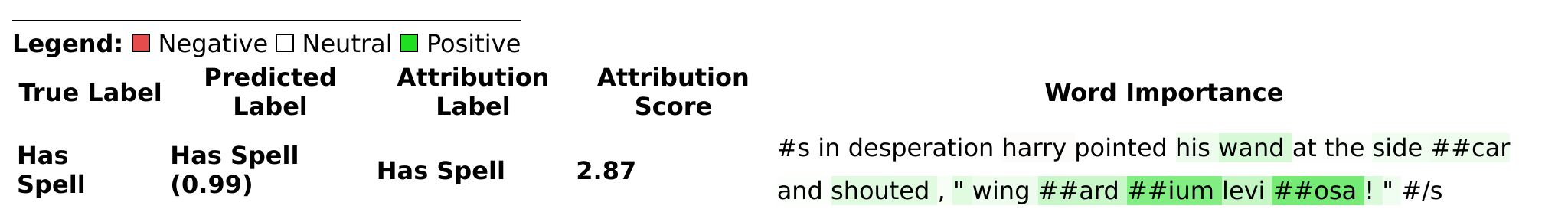}
  \caption{True positive classified sequences}
  \label{fig:interpret_study_sentence_split_true}
\end{figure}

\begin{figure}
  \includegraphics[width=\linewidth]{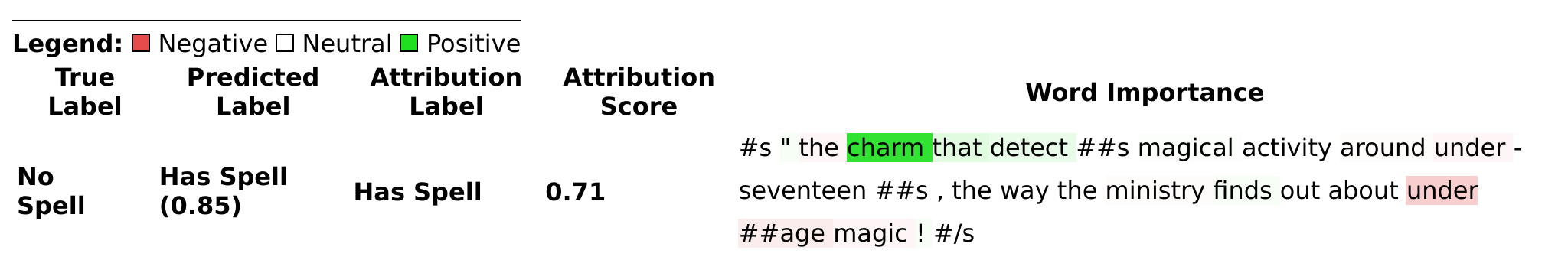}
  \includegraphics[width=\linewidth]{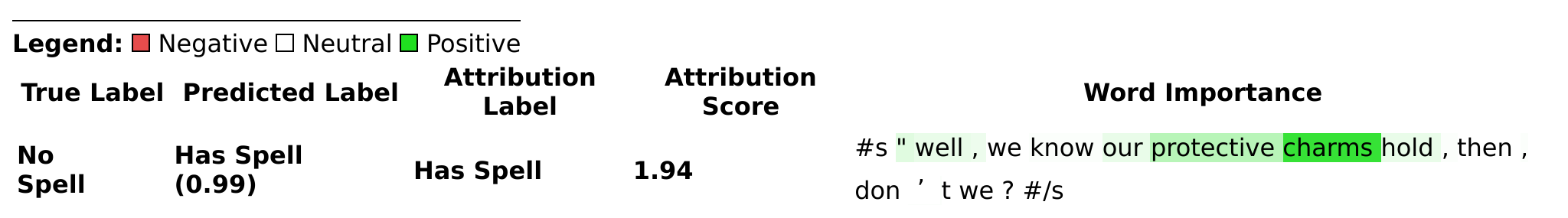}
  \caption{False positive classified sequences}
  \label{fig:interpret_study_sentence_split_false}
\end{figure}

Figure \ref{fig:interpret_study_sentence_split_false} shows as an example two of the false positive sequences analysed with the Transformers-Interpret library in order to identify some tokens that contribute particularly strongly to the sequence classification. The tokens that positively influence the result are marked in green and the opposite ones in red. The more intense the color, the higher the weighting. The attribute score is a summation of all weights, which individually fits in a range from -1 to 1 for a token. The attribute label indicates which label is being searched for. The predicted label is defined by calculating the weights of all tokens and normalising them to a range from 0 (negative) to 1 (positive).

The trained models generally do a good job of recognising sentences that correlate in some way with the subject matter of spells. However, some entries are still incorrectly classified as positive although they do not contain any of the phrases searched for, but terms that are often found in the context of a spell.
Therefore, the true context of a spell is not yet concretely recognised. The model still focus too much on single words, such as the word "charm". This may be because the model was trained on a too small dataset, or because the context is still too small.

\begin{figure}
  \includegraphics[width=\linewidth]{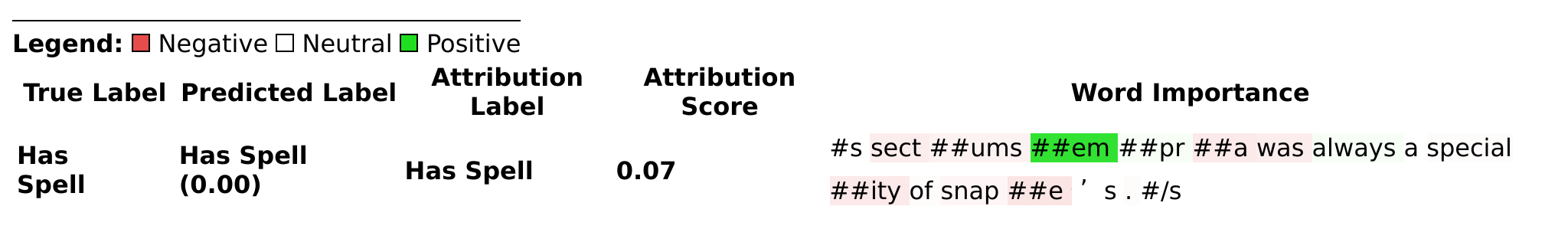}
  \caption{Tokenised spell "sectumsempra"}
  \label{fig:interpret_study_sentence_split_sectumsempra}
\end{figure}

When examining the individual tokens created by the tokenisation step, it was found that spells were more often split into several tokens than more common words. This is because the original texts used to create the tokeniser did not contain such terms. Based on the Wordpiece tokenisation approach, less common words are more often split into smaller tokens. 
This behaviour has sometimes resulted in some tokens of a word being classified as negative, but other tokens of the same word being strongly positive. This can be seen in Figure \ref{fig:interpret_study_sentence_split_sectumsempra} with the spell "sectumsempra". 

Therefore, before training the next model (1b), the Harry Potter spells are added to the tokeniser dictionary. This means that the spells are no longer divided into several tokens, and it is investigated whether the spells can be better recognised as a whole in order to improve the classification of the context. A total of 173 new tokens are added to the dictionary. The rest of the training remains the same.

\begin{table}[t]
\begin{subtable}{0.48\textwidth}
    \begin{small}
        \begin{tabular}{ c l | l l}
            \multicolumn{2}{l}{} & \multicolumn{2}{c}{Predicted} \\
            &  & \textbf{Positive} & \textbf{Negative} \\
        \cline{2-4}
            \multirow{2}{*}{\rotatebox[origin=tr]{90}{True}} & \textbf{Positive} & 278 TP & 6 FN \\
            & \textbf{Negative} & 75 FP & 13.883 TN \\[5pt]
            & \multicolumn{3}{l}{F1-Sore: 0,8728} \\
        \end{tabular} 
    \end{small}
    \caption{Big Dataset} 
\label{tab:study_sentence_tokenizer_combined}
\end{subtable}
\hspace*{0.02\textwidth} 
\begin{subtable}{0.48\textwidth}
    \begin{small}
        \begin{tabular}{ c l | l l}
            \multicolumn{2}{l}{} & \multicolumn{2}{c}{Predicted}  \\
            &  & \textbf{Positive} & \textbf{Negative} \\
        \cline{2-4}
            \multirow{2}{*}{\rotatebox[origin=tr]{90}{True}} & \textbf{Positive} & 111 TP & 1 FN \\
            & \textbf{Negative} & 42 FP & 14.088 TN \\[5pt]
            & \multicolumn{3}{l}{F1-Sore: 0,8377} \\
        \end{tabular} 
    \end{small}
    \caption{Incantation Dataset} 
    \label{tab:study_sentence_tokenizer_incantations}
\end{subtable}
\caption{Metrics of trained model 1b}
\label{tab:study_sentence_tokenizer_split}
\end{table}

\begin{figure}[h]
  \includegraphics[width=\linewidth]{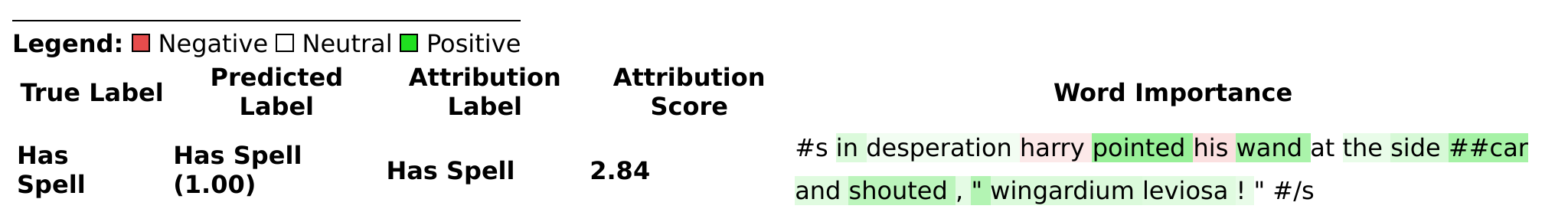}
  \caption{True positive classified sequences with advanced tokeniser}
  \label{fig:interpret_study_sentence_tokenizer_split}
\end{figure}

Updating the tokeniser had a positive effect on the F1 score. The exact confusion matrix of the results can be found in Table \ref{tab:study_sentence_tokenizer_split}. 
Compared to the (1a) model, the number of falsely correctly classified sequences has decreased. Figure \ref{fig:interpret_study_sentence_tokenizer_split} again shows the same sequence in comparison to figure \ref{fig:interpret_study_sentence_split_true}, where it can be seen that the words of the incantation "Wingardium Leviosa" are no longer divided into several tokens. It can also be seen that not only the incantation itself, but also some other words from the context are significant for the positive classification. 

\begin{figure}[h]
  \includegraphics[width=\linewidth]{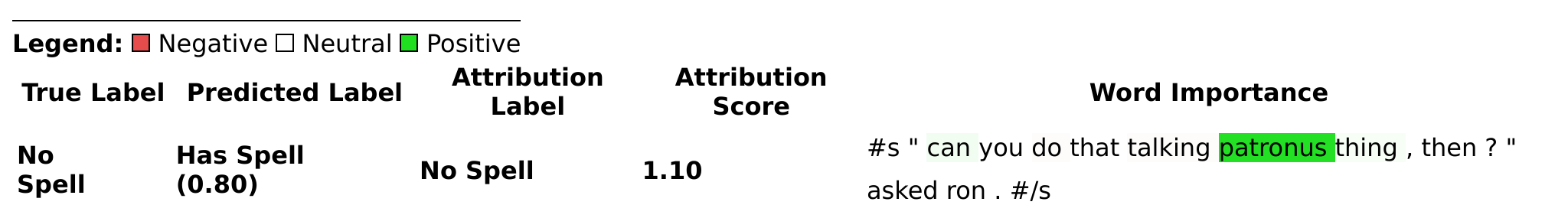}
  \caption{False positives classified sequences with advanced tokeniser}
  \label{fig:interpret_study_sentence_tokenizer_split_falsly}
\end{figure}

Through investigating the falsely positive classified sequences it is recognizable that some words are interpreted as false positives which are similar to a spell name, like the word "Patronus" seen in figure \ref{fig:interpret_study_sentence_tokenizer_split_falsly}.

\subsubsection{Paragraph-Split}
The models (2) were trained on datasets created by the Paragraph-Split to evaluate the impact of increasing the context scope from previously one sentence expanded to a full paragraph based on the corpus. The splitting approach reduces the number of entries in the training dataset; because one paragraph in the Harry Potter novels contains on average 1-3 sentences. 
By also using 10 times the proportion of negative as opposed to positive entries, the total size of the training dataset is 2,882 entries for the incantation-only dataset and 7,931 entries for the full dataset. The number of entries in the evaluation dataset is 6.911.

\begin{table}[t]
\begin{subtable}{0.48\textwidth}
    \begin{small}
        \begin{tabular}{ c l | l l}
            \multicolumn{2}{l}{} & \multicolumn{2}{c}{Predicted} \\
            &  & \textbf{Positive} & \textbf{Negative} \\
        \cline{2-4}
        \multirow{2}{*}{\rotatebox[origin=tr]{90}{True}} & \textbf{Positive} & 262 TP & 9 FN \\
        & \textbf{Negative} & 43 FP & 6.597 TN \\[5pt]
        & \multicolumn{3}{l}{F1-Sore: 0,9097} \\
        \end{tabular} 
    \end{small}
    \caption{Full Dataset} 
\label{tab:study_paragraph_combined}
\end{subtable}
\hspace*{0.02\textwidth} 
\begin{subtable}{0.48\textwidth}
    \begin{small}
        \begin{tabular}{ c l | l l}
            \multicolumn{2}{l}{} & \multicolumn{2}{c}{Predicted}  \\
            &  & \textbf{Positive} & \textbf{Negative} \\
        \cline{2-4}
            \multirow{2}{*}{\rotatebox[origin=tr]{90}{True}} & \textbf{Positive} & 106 TP & 2 FN \\
            & \textbf{Negative} & 23 FP & 6.780 TN \\[5pt]
            & \multicolumn{3}{l}{F1-Sore: 0,8945} \\
        \end{tabular} 
    \end{small}
    \caption{Incantation Dataset} 
    \label{tab:study_paragraph_incantations}
\end{subtable}
\caption{Metrics of trained model 2}
\label{tab:study_paragraph_split}
\end{table}

\begin{figure}[h]
  \includegraphics[width=\linewidth]{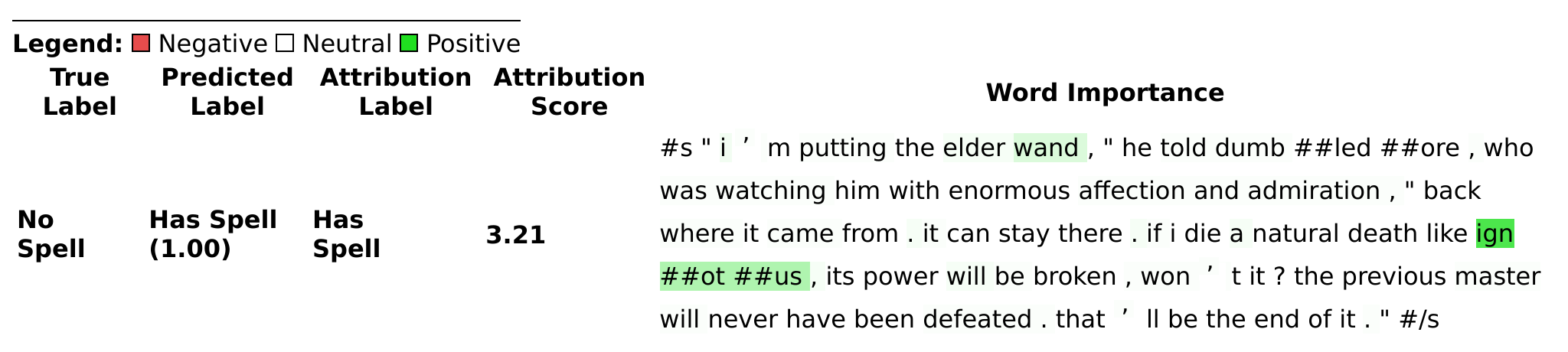}
  \caption{False positives classified sequences with the Paragraph-Split}
  \label{fig:interpret_study_paragraph_split_falsly}
\end{figure}

By increasing the average context size, an improvement over the previous models could be achieved. The confusion matrices of the tables 
\ref{tab:study_paragraph_split} give F1 scores of 0.8945 and 0.9097.
However, when analysing the sequences classified as false positives, it is noticeable that the models still pay a lot of attention to words like "charm", "spell", "charming" or "mark", which are words that are often used together with a spell or to form a spell name, but do not define a spell on their own. In addition, universe-specific character names such as "Regulus Arcturus Black", "Voldemort" or "Ignotus" are often recognised as spells, as shown in Figure   \ref{fig:interpret_study_paragraph_split_falsly}.

Despite the use of paragraphs, and therefore a reduced number of record entries, the number of sequences with a spell has only decreased slightly; because of their literary significance, a paragraph with a spell often consists of only one sentence, which ensures that a number of sequences with a spell are still quite short. Therefore, even though the results are quite good, there is still a slight problem that some contexts of sequences are not sufficient.

\subsubsection{Sequence-Split}
In order to train the model (3a) it was now investigated whether the context recognition could be better if the maximum possible processable length of a sequence is filled in as much as possible. For this purpose, the sentences are split by splitting the corpus using the Sequence-Split approach, where an entry contains the maximum number of concatenated sentences that the basic transformer model can process in a sequence.
This significantly reduces the number of possible sequences to 3,640 for the training dataset and 805 for the evaluation dataset. A sequence contains on average 3-5 sentences and is again labelled as positive if it contains an incantation or a spell name.
Due to the fact that the ratio of positive to negative entries increases significantly, it reduces the number of negative entries in the dataset, as the total number of possible sequences is less than the previously used ratio of 10 times negative to positive entries. 

\begin{table}[t]
\begin{subtable}{0.48\textwidth}
    \begin{small}
        \begin{tabular}{ c l | l l}
            \multicolumn{2}{l}{} & \multicolumn{2}{c}{Predicted} \\
            &  & \textbf{Positive} & \textbf{Negative} \\
        \cline{2-4}
        \multirow{2}{*}{\rotatebox[origin=tr]{90}{True}} & \textbf{Positive} & 197 TP & 7 FN \\
        & \textbf{Negative} & 11 FP & 587 TN \\[5pt]
        & \multicolumn{3}{l}{F1-Sore: 0,9563} \\
        \end{tabular} 
    \end{small}
    \caption{Full Dataset} 
\label{tab:study_sequence_combined}
\end{subtable}
\hspace*{0.02\textwidth} 
\begin{subtable}{0.48\textwidth}
    \begin{small}
        \begin{tabular}{ c l | l l}
            \multicolumn{2}{l}{} & \multicolumn{2}{c}{Predicted}  \\
            &  & \textbf{Positive} & \textbf{Negative} \\
        \cline{2-4}
        \multirow{2}{*}{\rotatebox[origin=tr]{90}{True}} & \textbf{Positive} & 77 TP & 7 FN \\
        & \textbf{Negative} & 5 FP & 716 TN \\[5pt]
        & \multicolumn{3}{l}{F1-Sore: 0,9277} \\
        \end{tabular} 
    \end{small}
    \caption{Incantation Dataset} 
    \label{tab:study_sequence_incantations}
\end{subtable}
\caption{Metrics of trained model 3a}
\label{tab:study_sequence_split}
\end{table}

\begin{figure}[h]
  \includegraphics[width=\linewidth]{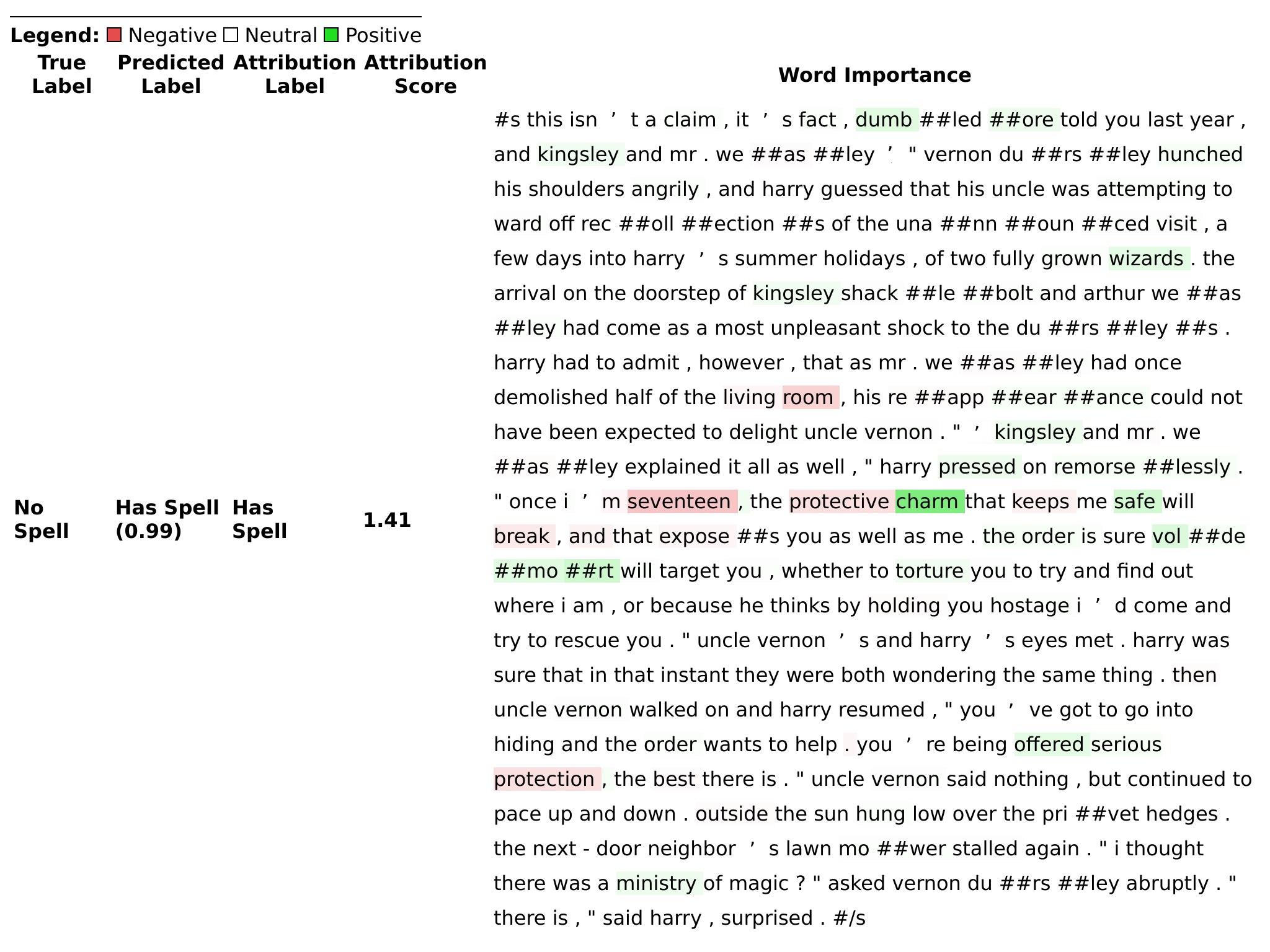}
  \caption{False positives classified sequences with the Sequence-Split}
  \label{fig:interpret_study_sequence_split_falsly}
\end{figure}

As shown in Table \ref{tab:study_sequence_split}, the model achieves a very satisfactory recognition rate with F1 scores of 0.9277 and 0.9563. It should be noted that the use of longer sequences and thus a larger context has a significant influence on the recognition rate of sequences with a Harry Potter spell for sequence classification. The single misclassified sequences seem to be outliers or very complex scenarios. 
The few falsely positively classified sequences can be characterised as containing individual terms such as "charm", "jinx" or " wizard" that are crucial for a positive classification, as seen in figure \ref{fig:interpret_study_sequence_split_falsly}. Such terms are often associated with a spell, or describe categories of spells such as "protective charms", but don't identify individual spells. Also very problematic are sentences in which a spell is not fully performed, and therefore these sentences were not identified as positive when the dataset was created, such as in the sentences \emph{"Piertotum - oh, for heaven's sake, Filch, not now [...]"} or \emph{“[...] he glimpsed another Death Eater swooping out of the way and heard, "Avada -"}, both of which contain only the beginning of a spell. This type of unfinished spell wasn't taken into account when creating the dataset. 

Due to the success in adapting the tokeniser to the corpus under investigation, the searched terms were again added to the tokeniser's dictionary as new tokens before training the model (3b) in order to adapt the embedding layer. The models are then trained on the same dataset using the sequence-split approach as before. 

The evaluation shows only a small change compared to the previous study without adapted tokeniser with F1 scores of 0.9697 and 0.9559, as shown in table 
\ref{tab:study_sequence_tokenizer_split}. Two other sequences were successfully classified. Instead, two other sequences were incorrectly classified as negative. The classification of all other sequences did not change.

\begin{table}[t]
\begin{subtable}{0.48\textwidth}
    \begin{small}
        \begin{tabular}{ c l | l l}
            \multicolumn{2}{l}{} & \multicolumn{2}{c}{Predicted} \\
            &  & \textbf{Positive} & \textbf{Negative} \\
        \cline{2-4}
        \multirow{2}{*}{\rotatebox[origin=tr]{90}{True}} & \textbf{Positive} & 195 TP & 9 FN \\
        & \textbf{Negative} & 9 FP & 589 TN \\[5pt]
        & \multicolumn{3}{l}{F1-Sore: 0,9559} \\
        \end{tabular} 
    \end{small}
    \caption{Full Dataset} 
\label{tab:study_sequence_tokenizer_combined}
\end{subtable}
\hspace*{0.02\textwidth} 
\begin{subtable}{0.48\textwidth}
    \begin{small}
        \begin{tabular}{ c l | l l}
            \multicolumn{2}{l}{} & \multicolumn{2}{c}{Predicted}  \\
            &  & \textbf{Positive} & \textbf{Negative} \\
        \cline{2-4}
        \multirow{2}{*}{\rotatebox[origin=tr]{90}{True}} & \textbf{Positive} & 80 TP & 3 FN \\
        & \textbf{Negative} & 2 FP & 717 TN \\[5pt]
        & \multicolumn{3}{l}{F1-Sore: 0,9697} \\
        \end{tabular} 
    \end{small}
    \caption{Incantation Dataset} 
    \label{tab:study_sequence_tokenizer_incantations}
\end{subtable}
\caption{Metrics of trained model 3b}
\label{tab:study_sequence_tokenizer_split}
\end{table}

\subsection{Token classification studies}
\label{chap:token_classification_studies}
\begin{figure}
\centering
  \includegraphics[width=\linewidth]{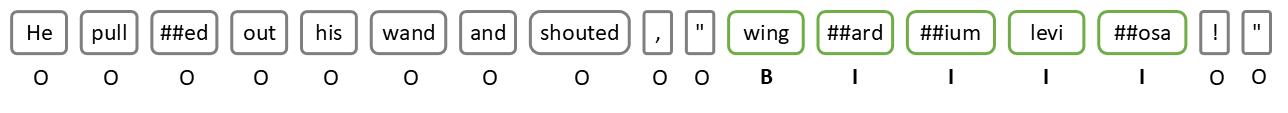}
  \caption{Token classification with the IOB scheme}{\begin{small}
  The illustration shows the phrase \emph{He pulled out his wand and shouted, "Wingardium Leviosa!"} in tokenized form. Below are the labels of each token in the IOB (Inside, Outside, Beginning) classification scheme.\end{small}}
  \label{fig:iob_schema}
\end{figure}

In addition to the sequence classification studies, token classification studies will be conducted. The aim is to find out whether it is possible to recognise not only the context, but also the spells explicitly at word level. Therefore, the basic model will be adapted with a different head layer for token classification, which will be trained to classify every single token of a sequence. 
The training procedure is basically the same as before, except that labels are now required for each token instead of one label for the whole sequence. 
Therefore, the dataset is created by labelling each token of a sequence with an IOB  (Inside, Outside, Beginning) scheme [109], where each individual spell token is given a respective “B” or “I” label and all others are given an “O” label. Figure \ref{fig:iob_schema} shows an example of how the tokens of a sentence labelled with the IOB scheme look like.
The output of the models is now a prediction of the labels for each token. The searched phrases are evaluated by "soft matching", where all words that have been split into several tokens are reassembled to make a word level statement. Split words can be recognised by the prefix \#\# for the following parts of the word. 
For comparison, the classification of a sequence is considered correct if at least one of the searched tokens is correctly classified with an I- or B-tag. However, if no spell token is tagged with an I- or B-tag, the sequence is classified as a false negative. If a spell is missing, the sequence is categorised as a false negative. This allows a metric similar to sequence classification to be created and compared.

\begin{table}[ht]
\centering
\begin{tabular}{ c | l | l }
    & \multicolumn{2}{c}{\textbf{F1-Score}} \\
    \# & Full & Inc  \\
    \cline{1-3}
    (1 a) & 0,7977 & 0,6769 \\
    (1 b) & 0,2880 & 0,3060 \\
\end{tabular}
\caption{Comparative overview of all token classification studies}
\label{tab:study_token_overview}
\end{table}

Table \ref{tab:study_token_overview} shows a comparative overview of the F1-score evaluation results for all trained token classification models.
For the models (1a) the dataset is also created by the Sentence-Split procedure. 
Furthermore, the dataset is mixed by using ten times the number of sequences without a spell for a positive sequence with a spell.
The models trained in this way achieve F1 scores of 0.6769 and 0.7977 respectively, which is slightly worse compared to the sequence classification approach. The corresponding metrics can be seen in Table \ref{tab:study_token_split}.
It is noteworthy that the false positive predicted sequences have mostly worsened. Often, punctuation marks such as "!", "." or "?", which are often placed immediately after an incantation in the corpus, are classified positively. 
However, words such as "charming", "gernumbli", "vunderful" or "victoire" were also incorrectly marked as positive.
This can happen because spells are usually broken down into several tokens, which makes it very likely that single tokens occur in spells as well as in parts of other words. Such a problem was already recognised in the previous research.

\begin{table}[t]
\begin{subtable}{0.48\textwidth}
    \begin{small}
        \begin{tabular}{ c l | l l}
            \multicolumn{2}{l}{} & \multicolumn{2}{c}{Predicted} \\
            &  & \textbf{Positive} & \textbf{Negative} \\
        \cline{2-4}
        \multirow{2}{*}{\rotatebox[origin=tr]{90}{True}} & 
        \textbf{Positive} & 274 TP & 10 FN \\
        & \textbf{Negative} & 129 FP & 13.829 TN \\[5pt]
        & \multicolumn{3}{l}{F1-Sore: 0,7977} \\
        \end{tabular} 
    \end{small}
    \caption{Full Dataset} 
\label{tab:study_token_combined}
\end{subtable}
\hspace*{0.02\textwidth} 
\begin{subtable}{0.48\textwidth}
    \begin{small}
        \begin{tabular}{ c l | l l}
            \multicolumn{2}{l}{} & \multicolumn{2}{c}{Predicted}  \\
            &  & \textbf{Positive} & \textbf{Negative} \\
        \cline{2-4}
            \multirow{2}{*}{\rotatebox[origin=tr]{90}{True}} & 
            \textbf{Positive} & 110 TP & 2 FN \\
            & \textbf{Negative} & 103 FP & 14.027 TN \\[5pt]
            & \multicolumn{3}{l}{F1-Sore: 0,6769} \\
        \end{tabular} 
    \end{small}
    \caption{Incantation Dataset} 
    \label{tab:study_token_incantations}
\end{subtable}
\caption{Metrics of trained models 1a}
\label{tab:study_token_split}
\end{table}

Therefore, before training the models (1b), the tokeniser is adapted by adding the spells from the corpus as new tokens. The remaining training and the dataset stays unchanged.
The results shown in Table \ref{tab:study_token_tokenizer_split} indicate that such an addition had a rather negative effect compared to sequence classification. This could be due to the use of a "soft matching" evaluation procedure, which makes it sufficient if only one part of a spell, i.e. one token out of several, can be classified positively. 
However, since the spells now consist of far fewer tokens than before, it is precisely these tokens that need to be positively classified. Nevertheless, it can be seen that the number of false positive sequences has also been greatly reduced by this adjustment. This confirms the assumption that some of the tokens that were previously classified as positive also occurred in other words, which is why they were falsely classified as positive. Even though in principle it can be assumed that the transformer can easily prevent this anomaly by its contextual reference, it seems that differently labelled subword parts still represent a complexity.

\begin{table}[ht!]
\centering
\begin{subtable}{0.48\textwidth}
    \begin{small}
        \begin{tabular}{ c l | l l}
            \multicolumn{2}{l}{} & \multicolumn{2}{c}{Predicted} \\
            &  & \textbf{Positive} & \textbf{Negative} \\
        \cline{2-4}
        \multirow{2}{*}{\rotatebox[origin=tr]{90}{True}} & 
        \textbf{Positive} & 54 TP & 230 FN \\
        & \textbf{Negative} & 57 FP & 13.901 TN \\[5pt]
        & \multicolumn{3}{l}{F1-Sore: 0,2880} \\
        \end{tabular} 
    \end{small}
    \caption{Full Dataset} 
\label{tab:study_token_tokenizer_combined}
\end{subtable}
\hspace*{0.02\textwidth} 
\begin{subtable}{0.48\textwidth}
    \begin{small}
        \begin{tabular}{ c l | l l}
            \multicolumn{2}{l}{} & \multicolumn{2}{c}{Predicted}  \\
            &  & \textbf{Positive} & \textbf{Negative} \\
        \cline{2-4}
            \multirow{2}{*}{\rotatebox[origin=tr]{90}{True}} & 
            \textbf{Positive} & 28 TP & 84 FN \\
            & \textbf{Negative} & 43 FP & 14.087 TN \\[5pt]
            & \multicolumn{3}{l}{F1-Sore: 0,3060} \\
        \end{tabular} 
    \end{small}
    \caption{Incantation Dataset} 
    \label{tab:study_token_tokenizer_incantations}
\end{subtable}
\caption{Metrics of trained models 1b}
\label{tab:study_token_tokenizer_split}
\end{table}

Adding the spells as new tokens to the tokeniser surprisingly contributed negatively to the token classification, contrary to the sequence classification.

In addition, some research has been done to analyse token classification with much larger sequences. However, in contrast to sequence classification, large sequences didn't give better results. 
One reason for this could be the small number of spells, which provide too few positively labelled tokens for a proper token classification approach. Theoretically, the model gives the best results when it simply classifies all tokens as negative.
Therefore, longer sequences with few positive tokens are rather unfavourable and shorter sequences tend to be more suitable for spell recognition using token classification. This is in contrast to sequence classification, where longer sequences were more successful.

\subsection{Universe comprehensive studies}
\label{chap:cross_universe_studies}
Finally, it will be investigated whether the trained models can also recognise the context of spells from other universes, such as the Eragon [35] or The Farewell Paladin [48] universes. This raises the question of whether features of spells can be generalised across universes, or whether they are too specific to individual universes.

Although there are spells in these universes, it was not possible to create a unique spell dataset for each universe in this elaboration due to certain characteristics of each universe. For example, in "Eragon", the phrases used as spells are also used for communication purposes, making them ambiguous and making it difficult to create an automated dataset. For "The Faraway Paladin" universe, no sources with a list of all spells from the universe have been found.
For these reasons, it is not possible to create an exact scoring dataset and therefore no unique metric can be given to classify the sequences. Nevertheless, the universes are useful for finding out which sequences are classified as positive or negative by each model.

The models (4a, 6, 8a) are used to classify sequences from the respective universe and to further analyse the influence of sequences of different sizes. For this purpose, the corpus of the foreign universes will be split for the study, using either the sentence-, paragraph- or sequence-split approach. The first book of each novel series is used as the corpus.
The results of the models used seem to be very interesting and can be seen in the table \ref{tab:study_cross_overview}. The table shows the number of sequences generated by each splitting approach, as well as the number of sequences positively classified by each model, for both universes. Each positively classified sequence is manually examined to see if it actually contains a spell, which is also shown in the table. However, due to the lack of information on sequences incorrectly classified as negative, no F1 score can be determined.

An unexpected result turned out that some spells from other universes could be recognised by the trained models. One overlapping criterion found across universes is that spells are often cast in the form of commands or exclamations. This can certainly be recognised by a Transformer model. However, there is also a discrepancy in the way spells appear in different fantasy universes. For example, some features of spells in the Harry Potter universe, such as the frequent use of the word "mark" or "spell" to form the wizard's name, have often led to the incorrect classification of sequences from other universes. It was also found that smaller sequences produced better results than the larger ones. The best results were obtained using the Paragraph-Split approach.
Nevertheless, the research showed that creating an overlapping dataset with spells from different universes could lead to promising results. Some spells from other universes have already been recognised by a model trained on only one universe.
However, the dataset can still be improved.

\begin{table}[h!] 
\centering
\begin{small}
    \begin{tabular}{ c l || l | l | l}
        & \textbf{Number} & \textbf{Sentence-Split} & \textbf{Paragraph-Split} & \textbf{Sequence-Split} \\
        \toprule
         \multirow{3}{*}{\rotatebox[origin=tr]{90}{\parbox{0.85cm}{\centering{Eragon}}}} 
         & \textbf{All } & 14.009 & 4.061 & 619 \\
         & \textbf{Positive} & 65 & 32 & 1 \\
         & \textbf{True Positive} & 24 & 18 & 0 \\
        \hline
         \multirow{3}{*}{\rotatebox[origin=tr]{90}{\parbox{1.4cm}{\centering{The Fareway Paladin}}}} 
         & \textbf{All } & 6.155 & 2.434 & 262 \\
         & \textbf{Positive} & 67 & 20 & 5 \\
         & \textbf{True Positive} & 25 & 18 & 2 \\
    \end{tabular}
\end{small}
\caption{Results of the investigations of alien universes} 
\label{tab:study_cross_overview}
\end{table}

\section{Summary \& Outlook}
\label{chap:summary}
In this study, several transformer-based models were trained to determine whether the context of casted magic spells from the fantasy novel Harry Potter could be efficiently detected . The obtained results of the investigations with a transformer approach fit into the result spectrum of other publications on sequence classification, such as the classification of the information content of Twitter tweets in the context of a coronavirus [115] or a general classification of the trustworthiness of information in tweets [116], each of which also achieved an F1 score above 0.9.

However, the lack of datasets brought some challenges for the definition of a magic spell dataset, such as different wordings, incomplete spell pronunciations, descriptive literary terms and several edge cases. In addition to incantations, which unambiguously define an actively cast spell, spell names can also be used as literary terms to describe the active execution of a spell. Descriptive terms such as "defensive spells" should also be considered spell names. Even if they don't explicitly name a spell, they are very similar to a spell name.

By investigating different context sizes by separating the corpus with different approaches to sequence classification, it was shown that the context of a single sentence or a small paragraph is often not sufficient to satisfactorily determine whether an active spell is being performed or not. However, when larger sequences of several sentences are used, a transformer-based model is quite efficient at recognising the context of a spell. Our research shows that the best results can be achieved by using the largest possible sequence length that can be processed by the model. However, it should be noted that larger sequences may lose some of the more detailed aspects.
In additional token classification studies, the initial model was trained more precisely to classify each individual token in a sequence, thus directly identifying the spells in a sequence. The results showed that shorter sequences were more successful, identifying individual spells much more frequently than longer sequences, contradicting our results from the sequence classification studies.
Further experiments were carried out to see if the models trained on the Harry Potter corpus could be used to identify the context of spells in other universes. 
The question arose as to whether the properties of spells could be generalised across multiple universes, or whether they were too specific to individual universes.
The most common feature of spells from different universes is the use of an imperative or pronunciation as an exclamation.
But it has also been shown that the use of a model trained only on spells from one universe lags behind the identification of characteristics of words for specific universes. For example, the use of the word \"spell\" to form a spell name in Harry Potter often leads to incorrect classifications of sequences from other universes where this phrase isn't used.

However, it is still possible to extend the results by making adjustments to the dataset and training. The trained model of this study has succeeded in the approach of clearly recognising the context of a spell. But, the model recognises not only sequences containing the searched phrases, but also some sentences that are completely out of context, as well as many sentences that are only vaguely about magic, but in which no spell is explicitly named or performed.

In conclusion, the use of a transformer to automatically classify sequences based on searched phrases within a context has produced very satisfactory results. Provided that authors adhere to a formal definition of magic and spells they have created, it is possible to identify spells within a universe and classify them with a high degree of accuracy using a Transformer model.
However, spells have always been a construct of the human imagination, making it difficult to define global formal rules for them. In particular, writers and inventors of fantasy universes have been and are free to interpret their definition of magic as they wish. This makes it difficult to create a universe-wide data set of magic spells, since this topic is handled differently by authors in different universes.
Even if many of the worlds, ideas and plots of fantasy novels are already part of academic discourses dealing with linguistic subtleties, neologisms or other aspects of fictional languages  \cite{analysingSpells, linguaculturalRole, namesAndNaming}, but also with ethnic, social or cultural issues \cite{rothfuss, nikolajeva, lawsAndFunctions} deal with specific individual fantasy universes, further research in the area of this paper would be supported by additional linguistic considerations of spells from different fantasy universes. 
However, based on the knowledge gained, it seems possible to create a comprehensive dataset of spells from different universes for post-training a Transformer model, which could certainly lead to promising results for efficient recognition of cross-universe spells.
\printbibliography[heading=bibintoc]

\section{Appendix}
\label{app:appendix}

\subsection{Harry Potter Spells}
The following table \ref{tab:incantations_list} lists all actively cast spells (incantations) together with their spell names that appear in the seven Harry Potter novels. The table \ref{tab:names_list} extends this list with spells which appear only by their spell names.

\begin{longtable}{| l p{1.2in} || l p{1.1in} |}
\caption{Incantations with Spell Names}
\label{tab:incantations_list}
\\ \hline
    \textbf{Incantation} & \textbf{Name} & \textbf{Incantation} & \textbf{Name} \\
\hline
\endfirsthead
\caption{\textit{(Continued)} Incantations with Spell Names} \\
\hline
    \textbf{Incantation} & \textbf{Name} & \textbf{Incantation} & \textbf{Name} \\
\hline
\endhead
\hline
    \multicolumn{4}{r}{\textit{Continued on the next page}} \\
\endfoot
\hline
\endlastfoot  
Accio & Summoning Charm & Locomotor Mortis & Leg-Locker Curse \\
Aguamenti & Aguamenti Charm & Lumos &  \\
Alohomora & Alohomora Charm & Meteolojinx Recanto &  \\
Anapneo &  & Mobiliarbus &  \\
Aparecium &  & Mobilicorpus &  \\
Avada Kedavra & Killing Curse & Morsmordre & Dark Mark \\
Cave Inimicum &  & Muffliato & Muffliato Spell, Muffliato Charm \\
Colloportus &  & Nox &  \\
Confundo & Confundus Charm & Obliviate & Memory Charm \\
Confringo & Blasting Curse & Obscuro &  \\
Crucio & Cruciatus Curse & Oppugno &  \\
Defodio & Gouging Spell & Orchideous &  \\
Deletrius &  & Peskipiksi Pesternomi &  \\
Densaugeo &  & Petrificus Totalus & Body-Bind Curse \\
Deprimo &  & Piertotum Locomotor & Piertotum \\
Descendo &  & Point Me & Four-Point Spell \\
Diffindo & Severing Charm & Portus &  \\
Dissendium &  & Prior Incantato & Reverse Spell \\
Duro &  & Priori Incantatem &  \\
Engorgio & Engorgement Charm & Protego & Shield Charm \\
Episkey &  & Protego Horribilis &  \\
Evanesco & Vanishing Spell & Protego Totalum &  \\
Expecto Patronum & Patronus Charm & Quietus &  \\
Expelliarmus & Disarming Charm, Disarming Spell & Reducio &  \\
Expulso &  & Reducto & Reductor Curse \\
Ferula &  & Relashio & Revulsion Jinx \\
Finite Incantatem &  & Rennervate &  \\
Finite &  & Reparo & \\ 
Flagrate &  & Repello Muggletum & Muggle-Repelling Charm \\
Furnunculus & Furnunculus Curse & Rictusempra & Tickling Charm \\
Geminio &  & Riddikulus & \\ 
Glisseo &  & Salvio Hexia & \\ 
Homenum Revelio &  & Scourgify & Scouring Charm \\
Impedimenta & Impediment Jinx, Impediment Curse & Sectumsempra & Sectumsempra Spell \\
Imperio & Imperius Curse & Serpensortia & \\ 
Impervius & Impervius Charm & Silencio & Silencing Charm \\
Incarcerous &  & Sonorus & \\ 
Incendio &  & Specialis Revelio &  \\
Langlock & Tongue-Tying Curse & Stupefy & Stunning Spell, Stunner \\
Legilimens & Legilimens, Legilimency & Tarantallegra &  \\
Levicorpus &  & Tergeo &  \\
Liberacorpus &  & Waddiwasi &  \\
Locomotor & Locomotion Charm & Wingardium Leviosa & Levitation Charm \\
\bottomrule
\end{longtable}

\hspace{0pt}

\begin{table}[t!]
\caption{Spell Names without Incantations}
\label{tab:names_list}
\centering
\begin{tabular}{| l || l || l |}
\hline 
    \multicolumn{3}{|l|}{\textbf{Name}} \\
    \toprule  
    Age Line & Entrail-Expelling Curse & Invisibility Spell \\
    Anti-Cheating Spell & Extension Charm & Jelly-Legs Jinx \\
    Anti-intruder jinx & Fidelius Charm & Occlumency \\
    Apparition, Apparate & Fiendfyre & Protean Charm \\
    Atmospheric Charm & Flagrante Curse & Refilling Charm \\
    Babbling Curse & Flame-Freezing Charm & Shock Spell \\
    Bat-Bogey Hex & Flying charm & Slug-vomiting Charm \\
    Bedazzling Hex & Freezing Charm & Stealth Sensoring Spell \\
    Bubble-Head Charm & Growth Charm & Sticking Charm \\
    Caterwauling Charm & Hair-thickening Charm & Stinging Hex, Stinging Jinx \\
    Cheering Charm & Homorphus Charm & Stretching Jinx \\
    Conjunctivitis Curse & Hover Charm & Supersensory Charm \\
    Cushioning Charm & Hurling Hex & Transmogrifian Torture \\
    Disapparition, Disapparate & Imperturbable Charm & Trip Jinx \\
    Disillusionment Charm & Inanimatus Conjurus Spell & Unbreakable Charm \\
    Drought Charm & Intruder Charm & Unbreakable Vow \\
    Enlargement Charm & &  \\
    \bottomrule
\end{tabular}
\end{table}
\end{document}